\documentclass{article}

\usepackage[numbers, compress]{natbib}

\usepackage[final]{neurips_2024}

\usepackage{array}
\usepackage{graphicx}
\usepackage[utf8]{inputenc} 
\usepackage[T1]{fontenc}    
\usepackage{hyperref}       
\usepackage{url}            
\usepackage{booktabs}       
\usepackage{amsfonts}       
\usepackage{nicefrac}       
\usepackage{microtype}      
\usepackage{xcolor}         
\usepackage{amsmath}
\usepackage{wrapfig,lipsum,booktabs}

\title{Generating Intermediate Representations for Compositional Text-To-Image Generation}

\author{
  Ran Galun \\
  The Hebrew University of Jerusalem\\
  \texttt{ran.galun@mail.huji.ac.il} \\
  \And
  Sagie Benaim \\
  The Hebrew University of Jerusalem\\
  \texttt{sagie.benaim@mail.huji.ac.il} \\
}

\begin{document}

\maketitle

\begin{abstract}
Text-to-image diffusion models have demonstrated an impressive ability to produce high-quality outputs. However, they often struggle to accurately follow fine-grained spatial information in an input text. To this end, we propose a compositional approach for text-to-image generation based on two stages. In the first stage, we design a diffusion-based generative model to produce one or more aligned intermediate representations (such as depth or segmentation maps) conditioned on text. In the second stage, we map these representations, together with the text, to the final output image using a separate diffusion-based generative model. 
Our findings indicate that such compositional approach can improve image generation, resulting in a notable improvement in FID score and a comparable CLIP score, when compared to the standard non-compositional baseline. Our code is available at \url{https://github.com/RANG1991/Public-Intermediate-Semantics-For-Generation}
\end{abstract}

\section{Introduction}
Recently, text-to-image generation has shown highly impressive results, primarily using diffusion modeling. To enable effective conditioning, one often integrates textual embeddings as input to the denoising network
~\cite{ramesh2021zero, ramesh2022hierarchical, nichol2021glide, rombach2022high}. However, text prompts may fail to enable full and precise spatial control. Describing semantic properties, such as the segmentation of individual objects, or physical characteristics, such as the depth of objects within an image, using solely textual descriptions, can be challenging and inefficient. Consequently, text-to-image diffusion models often need to implicitly infer these properties from the text-image data used during training. This process is prone to inaccuracies and may lead to difficulties in accurately capturing intermediate representations, such as object segmentation or depth. In this study, therefore, we ask whether a compositional approach of first explicitly generating these intermediate representations and subsequently using them as an additional condition can help mitigate these issues and improve text-to-image generation.

Some recent work considered the ability to condition the generation process on some intermediate representation in addition to text~\cite{zhang2023adding, Mou_Wang_Xie_Wu_Zhang_Qi_Shan_2024, jiang2024scedit, zheng2023layoutdiffusion}. However, these approaches require having existing intermediate representations (such as a segmentation map or a depth map) as input, which is not the case in text-to-image generation.
Another approach is to first generate an intermediate layout or blobs representation from an input text, and then generate the entire image using the layout or blobs representation and the input text. 
This can be done by using cross-attention \cite{lian2023llm, phung2024grounded, nie2024compositional}, or a separate diffusion model \cite{zhang2024realcompo}. 
However, since both a layout and a blob cannot encompass fine details, these approaches provide only partial control over the output image.

Instead, we propose to employ a two-step compositional process: (1). First, we generate fine-grained intermediate representations (such as a depth map or a segmentation map) conditioned on the input text. To do so, we fine-tune a pre-trained Stable-Diffusion~\cite{rombach2022high} model to generate an intermediate representation given a text prompt. In the case where more than one representation is generated, we propose an approach to align those representations, making sure that they correspond to the same output image. (2). We then generate the output image conditioned on the input text and the intermediate representations generated. We do so using a pre-trained ControlNet~\cite{zhang2023adding}, which was trained to generate an output image given both the text and the generated intermediate representations.

We consider three distinct intermediate representations for our experiments: a depth map, a segmentation map, and Hough lines (HED). We assess their impact on text-to-image generation. Our findings indicate that, among these single representations, using either the depth map or the segmentation map (solely, without alignment) as intermediate representations results in a notable improvement in Fréchet Inception Distance~\cite{heusel2017gans} (FID) score compared to standard non-compositional Stable-Diffusion baseline. In addition, we explore the generation capabilities using two aligned intermediate representations, revealing insights into their effectiveness and potential benefits. 

\section{Related Work}
\noindent \textbf{Conditional Text-to-Image Generation} \quad
Early advances were dominated by Generative Adversarial Networks (GANs)~\cite{mansimov2015generating, reed2016generative, xu2018attngan} 
and later approaches considered an autoregressive approach~\cite{ramesh2021zero, ding2021cogview, yu2022scaling}.
Recently, diffusion models \cite{ho2020denoising, dhariwal2021diffusion, nichol2021glide, saharia2022photorealistic} have achieved significant improvements. 
To enable conditional generation, Stable-Diffusion \cite{rombach2022high} incorporates conditions (e.g. a text prompt) by first encoding them and then applying cross-attention with the denoiser's layers. 
At inference time, classifier guidance~\cite{dhariwal2021diffusion} can be used to guide the noise trajectory using an external classifier. Alternatively, classifier free guidance~\cite{ho2022classifier}, combines the output of an conditional and unconditional model.
ControlNet~\cite{zhang2023adding} introduces replicated U-Net layers that share weights with the original Stable-Diffusion backbone U-Net. These replicated U-Net layers get as input a control image (e.g. a segmentation map).
\noindent \textbf{Compositional Text-to-Image Generation} \quad
Several studies considered text-to-image generation as a compositional approach, 
where first, the condition is generated, and only then the final image is generated based on this condition. Approaches used mainly Large-Language-Models (LLMs) to generate an intermediate representation (e.g., layout, blob)~\cite{lian2023llm, zhang2024realcompo, nie2024compositional}, but these do not enable fine-grained control. \cite{wang2024compositional} used a bounding box representation instead. 
\cite{luo2024readout} extracts readouts from intermediate features and guides the generation process based on a user input and the readouts. However, these methods still require user input and otherwise cannot achieve fine-grained control. 

\section{Method}
\begin{figure}[t!]
\centering
\begin{tabular}{cc}
\includegraphics[trim={0 2cm 0 0},clip,width=0.46\linewidth]{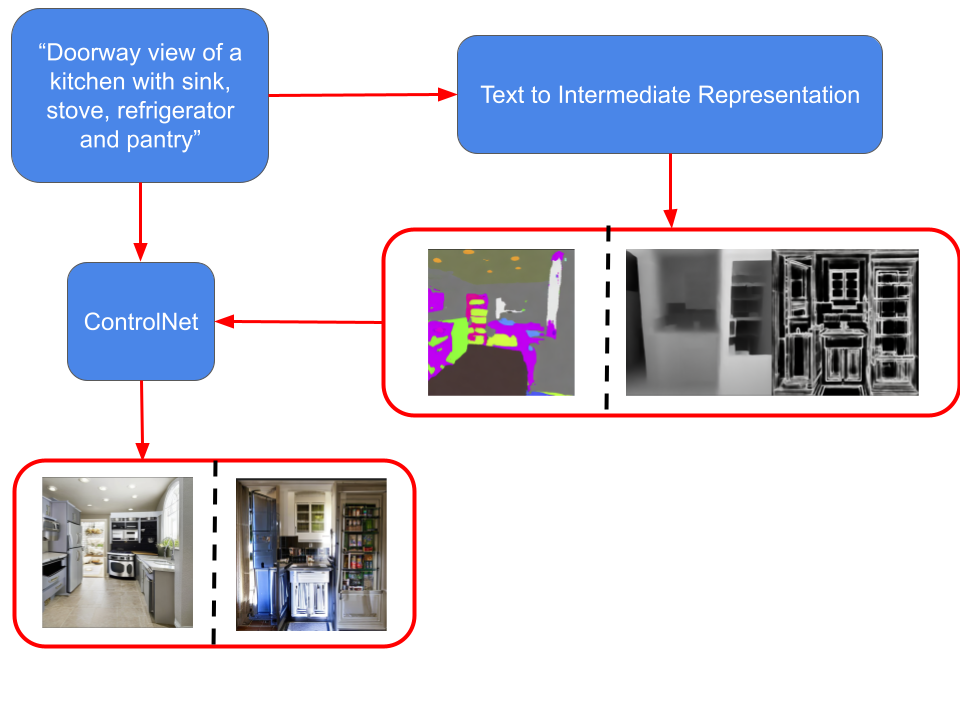} &  
\includegraphics[width=0.43\linewidth]{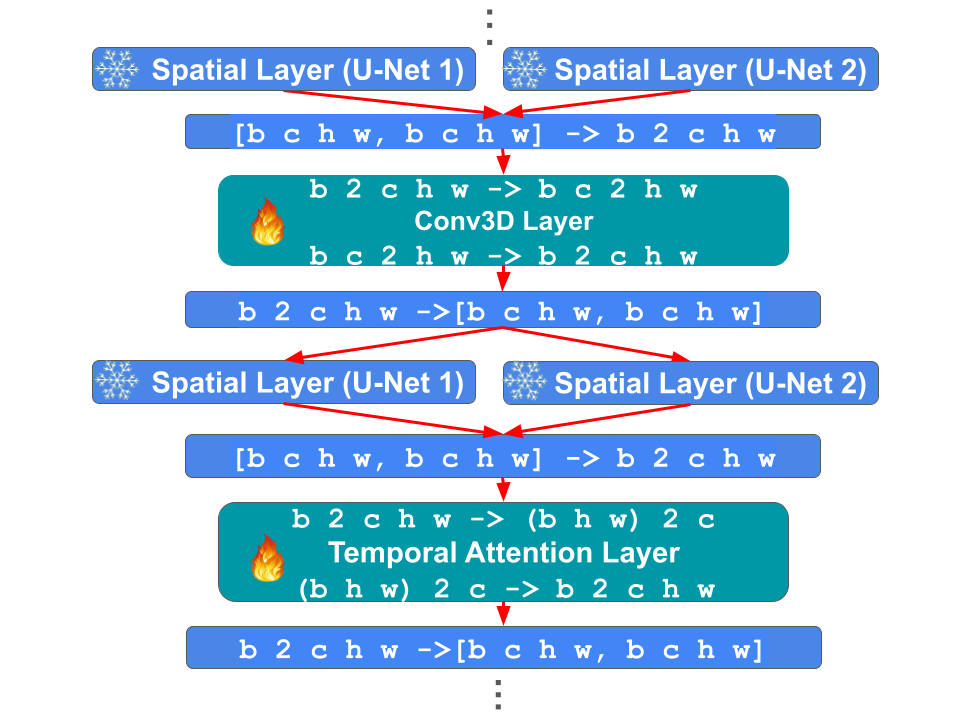}\\
(a) & (b) \\
\end{tabular}
\caption {(a). \textbf{Illustration of the full pipeline}. In the first step, we generate aligned intermediate representation(s) given the input text. In the second stage, we use a pre-trained ControlNet to map the input text and the generated intermediate representation(s) to an output image. 
(b). \textbf{Illustration of our alignment procedure}. Given two pre-trained text-to-intermediate models (e.g., text-to-depth and text-to-segmentation), we interleave their spatial layers using "temporal" layers. The "temporal" layers consist of either a 3D convolution or a temporal attention layer and indicate the dimension on which the attention or convolution is performed. For clarity, we also provide each component's input and output dimensions. We note that only the temporal layers are trained in this stage.  
}
\label{fig:illustration}
\vspace{-0.8cm}
\end{figure}

We describe here our two-step compositional generation approach, as illustrated in Fig.~\ref{fig:illustration}. 

\subsection{Generating Intermediate Representations}
To generate an intermediate representation given an input text, we consider a text-representation pairs dataset and fine-tune a text-to-image pre-trained Stable-Diffusion (SD) model on this dataset. 
As the SD's VAE was trained on images, we also fine-tune it on the intermediate representation.

\noindent \textbf{Aligned Intermediate Representations} \quad 
Our fine-tuned SD models can now be used to generate multiple intermediate representations. There is, however, a problem, since these different intermediate representations may not be aligned.
While we describe the alignment procedure of two intermediate representations here, this can be extended to a variable number of such representations. 
Our alignment procedure is inspired by that of \cite{blattmann2023align}, which tackles a different problem of aligning the latents of a text-to-image diffusion model to enable text-to-video generation.
In particular, we consider two unaligned pre-trained models, one for each intermediate representation.  
We assume Stable-Diffusion v2.1 model and denote the spatial layers within the U-Net (in both the encoding and decoding paths) of each model as $l_{\Theta_1}^i$ and $l_{\Theta_2}^i$ respectively, where $i$ refers to the layer index.


We now introduce a joint temporal layer, $l_\Phi^i$, between consecutive spatial layers. We assume $z^i_{crl_r} \in \mathbb{R}^{B \times C \times H \times W}$ is the output of $l_{\Theta_r}^i$ ($r=1,2$), 
where $C$ represents the number of latent channels, and $H$ and $W$ denote the spatial latent dimensions. 
We then concatenate $z^i_{crl_1}$ and $z^i_{crl_2}$ along a new dimension $t$, referred to as the "temporal" dimension ($t = 2$), resulting in $z^i_{crl}$.
$z^i_{crl}$ passes through two types of temporal mixing blocks: (i) temporal attention layers and (ii) residual blocks employing 3D convolutions. For (i), we used a block denoted as $f_{cross-attn}(z^i_{crl},c)$, that is defined as follows: 
\begin{wraptable}{r}{6.4cm}
\centering
\vspace{0.2cm}
\centering
\begin{tabular}{ccc}
\toprule
\centering
Method& $\downarrow$ FID& $\uparrow$ CLIP\\
\midrule
SD v2.1 (Baseline) & 23.44& \textbf{30.58}\\
Ours (Seg) & \textbf{19.92} & 30.43\\
Ours (Depth) & 20.73& 30.30\\
Ours (HED) & 27.56& 29.70\\
Ours (Depth \& HED)
& 50.56& 28.88\\
Ours (Depth \& Seg)
& 32.53& 29.82\\
\midrule
ControlNet - GT Seg & 16.80& 30.42\\
ControlNet - GT Depth & 16.17& 30.35\\
\bottomrule
\end{tabular}
\centering
\caption{FID and CLIP alignment scores. In brackets we note the type of the intermediate representation(s). The bottom two rows provide a comparison when the second stage is used with ground truth intermediate representation. While this is not a direct comparison (as it uses additional input), it provides an upper bound.}
\label{tab:num_FID_CLIP_results}
\vspace{-2.6cm}
\end{wraptable}
\begin{gather*}
f_{cross-attn-1} = \text{cross-attn}(\text{lin}_1(z^i_{crl}), \text{lin}_2(c)) \\
f_{l-norm-1} = \text{l-norm}(\text{lin}_1(z^i_{crl}) + f_{cross-attn-1}) \\
f_{cross-attn-2} = \text{cross-attn}(f_{l-norm-1}, \text{lin}_2(c)) \\
f_{l-norm-2} = \text{l-norm}(f_{l-norm-1} + f_{cross-attn-2}) \\
f_{cross-attn}(z^i_{crl},c)=\text{lin}_3(f_{l-norm-2})
\end{gather*}
where $c \in \mathbb{R}^{1 \times 1024}$ is the text CLIP embedding. 
$\text{lin}_i$'s are linear projection layers and $\text{l-norm}$ is a layer norm. 
Cross-attention is applied as in \cite{vaswani2017attention} between a batch of $(B \cdot H \cdot W)$ vectors with a sequence length of $2$ or $1$. Specifically, the queries in the cross-attention computation are the projected spatial outputs of the two U-Nets, and the keys and values are the projected text embeddings.

For (ii), we used the following block: $f_{conv}(z^i_{crl}) = \text{ReLU}(z^i_{crl} + \text{Conv3D}(z^i_{crl}))$. 
The input to the 3D convolution block is of shape $B \times C \times 2 \times H \times W$, and the output is of shape $B \times C \times 2 \times H \times W$. That is, we apply 3D convolution on $B$ 4-dimensional tensors. For the convolution parameters, we used a kernel size of $(3, 1, 1)$ and a stride of $1$.

Following either temporal mixing blocks (i) or (ii), we apply a residual operation: $\alpha_i \cdot z^i_{crl} + (1-\alpha_i) \cdot f$, where $\alpha_i = sigmoid (x)$ is a learned value between 0 and 1, and $f$ is either $f_{att}(z^i_{crl},c)$ or $f_{conv}(z^i_{crl})$. 
The temporal blocks are trained using standard SD reconstruction objectives on the output representations, given an input text. 

\subsection{Generating the final output image}
Given the previous stage, we can now generate a set of aligned intermediate representations conditioned on the input text. The second stage involves training a ControlNet~\cite{zhang2023adding} on a dataset of representation(s)-image pairs. To obtain such pairs we apply off the shelf method for obtaining such representations from images (such as depth estimation or segmentation). We then follow ControlNet's procedure of first training a ControlNet model for each intermediate representation in isolation and then combining them together to enable conditioning by all intermediate representations. The underlying SD model is still conditioned on text in the standard manner as in SD.  Once these models are trained, we simply generate the output image by using our generated intermediate representations and input text as a condition to the pre-trained ControlNet model. 

\section{Experiments}

\noindent \textbf{Implementation Details and Datasets} \quad
For the training phase, we fine-tuned SD v2.1 models on the intermediate representations (i.e. Depth map, Segmentation map, Hough lines) extracted from the first $300,000$ samples from MS-COCO \cite{lin2014microsoft} 2017 training set. We used Uniformer \cite{li2023uniformer} for segmentation estimation, ControlNet \cite{zhang2023adding} implementation for HED generation and Depth Anything \cite{yang2024depth} for depth estimation. We trained each of our models for 12 epochs on $40$GB GPUs, with a learning rate of $1e-5$ and a batch size of $32$. We used AdamW optimizer with a weight decay of $0.01$. For the evaluation phase, we followed previous papers and evaluated our models on $25,000$ samples from MS-COCO 2017 validation set. For the sampling process, we used 80 DDIM steps.

\begin{figure}[t!]
\begin{tabular}{cc}
\hspace{-0.5cm}
\includegraphics[trim={1cm 0.5cm 0cm 0cm},clip,width=0.469\linewidth]{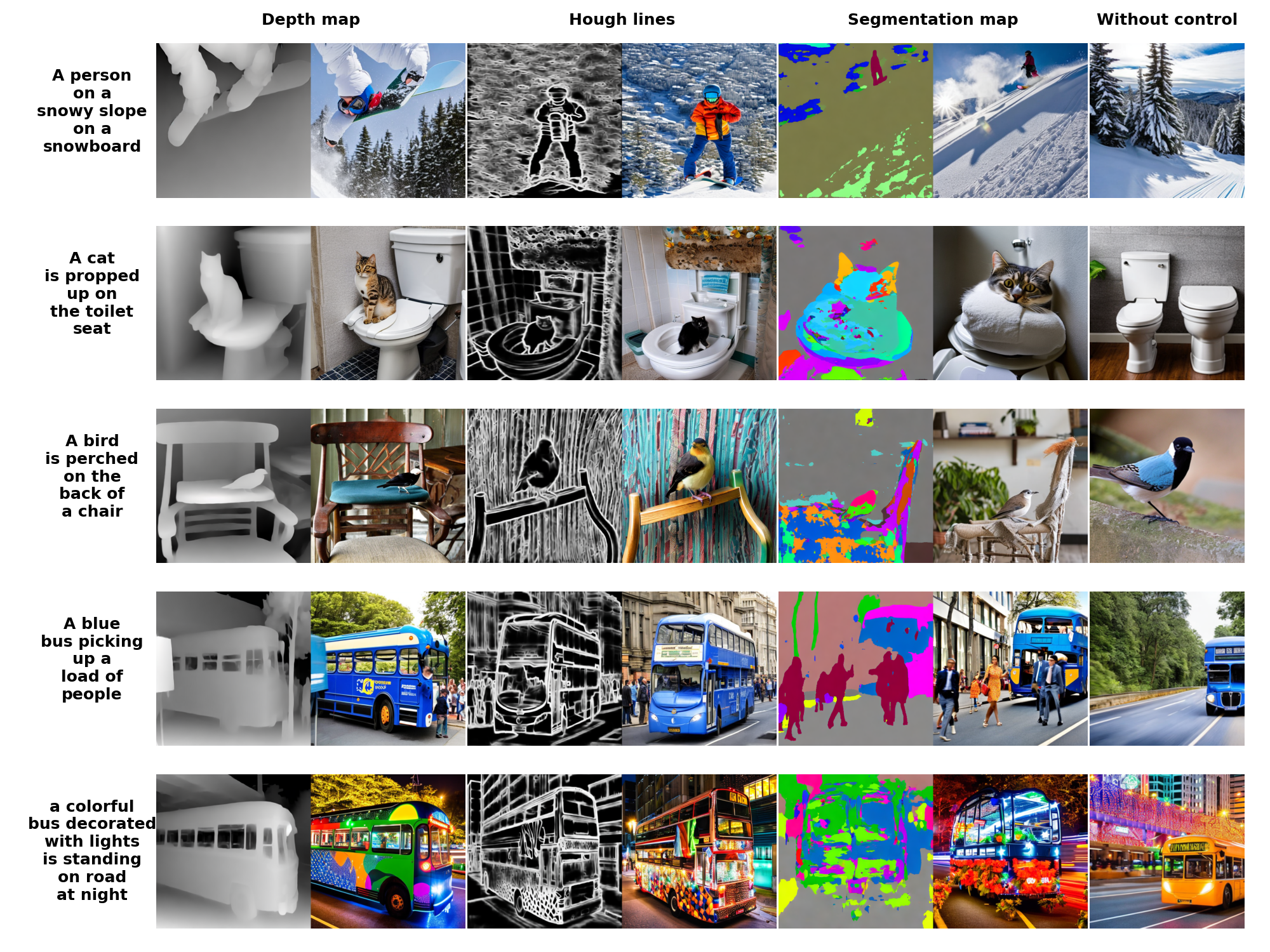} & 
\hspace{-0.4cm}
\includegraphics[trim={1cm 0.4cm 0cm 0},clip,width=0.531\linewidth]{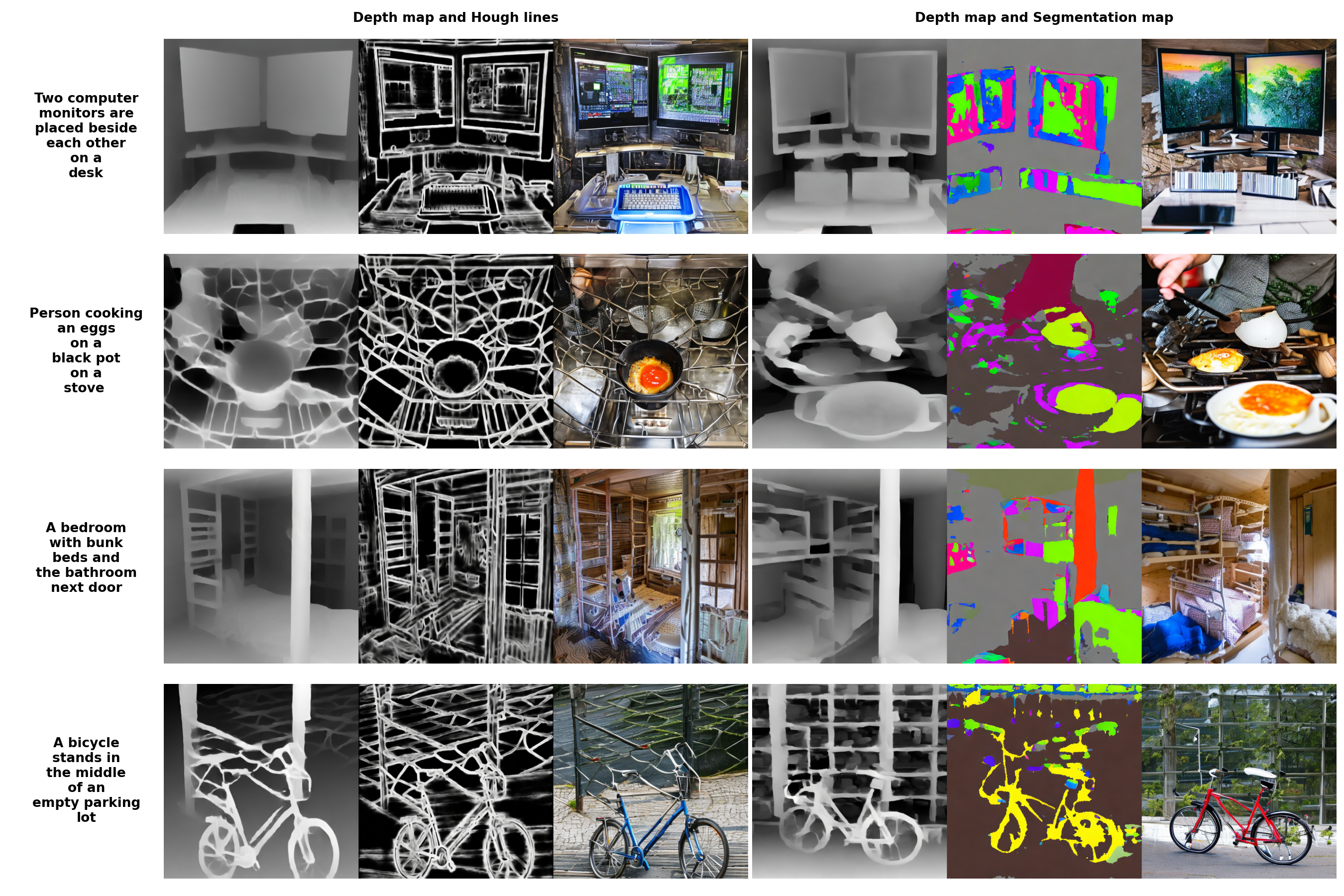} \\ 
(a) & (b) \\
\end{tabular}
\caption{(a). Results using a single intermediate representation (first six columns) and from original SD (last column). The generated intermediate representation is to the left of each output image. (b). As in (a), but using our aligned intermediate representations (depth \& HED or depth \& segmentation).} 
\label{im_FID_CLIP_results}
\vspace{-0.4cm}
\end{figure}

\noindent \textbf{Using A Single Intermediate Representations} \quad
Tab.\ref{tab:num_FID_CLIP_results} provides a numerical evaluation of FID and CLIP similarity scores in comparison to SD baseline. Our compositional approach achieves lower FID scores than the original non-compositional SD model, for both the depth and segmentation intermediate representations, except for HED. 
We hypothesize that this is a result of the domain shift that occurs between the generated Hough lines in contrast to the real Hough lines. As the second stage trains on such realistic data, this domain shift can result in significant errors. 
Fig.~\ref{im_FID_CLIP_results}(a), provides a visual illustration of outputs of our model in comparison to the SD baseline, showing examples whereby predicting intermediate representation improves overall text-image correspondence. 

\noindent \textbf{Using Aligned Intermediate Representations} \quad
Tab.~\ref{tab:num_FID_CLIP_results} and Fig.~\ref{im_FID_CLIP_results}(b)  present the corresponding numerical and visual results when using aligned intermediate representations. While our method produces aligned outputs, we observe a drop in performance, both in terms of FID and CLIP scores. We hypothesize that, while our alignment model produces aligned outputs, this  comes at the expense of quality. This quality degradation results in inputs which are far from real intermediate representation, resulting in a domain shift, which ultimately results in worse performance in the second stage when these intermediate representations are fed into ControlNet.


\section{Conclusion}

In this work, we proposed a two-stage compositional approach for text-to-image generation, comprising of first generating intermediate representations and subsequently using these representations to generate a final output image. Our compositional apporach demonstrated improved FID scores over the non-compositional baseline when using a single depth or segmentation maps as intermeidate representations. Future work could focus on refining the alignment process, and addressing the domain shift that occurs between generated representations and those used as input in the second stage of our compositional approach.

\medskip

\bibliographystyle{unsrtnat}
\bibliography{neurips_2024}


\appendix




\end{document}